\documentclass[10pt,twocolumn,letterpaper]{article}

\usepackage[pagenumbers]{cvpr} 

\usepackage{adjustbox}
\usepackage[table,xcdraw]{xcolor}

\usepackage{xcolor}

\renewcommand{\vec}[1]{\boldsymbol{#1}}
\newcommand{\matr}[1]{\boldsymbol{#1}}
\newcommand{\liegroup}[1]{\mathrm{#1}}
\newcommand{\liealgebra}[1]{\mathfrak{#1}}

\DeclareMathOperator*{\argmax}{arg\,max}

\newcommand{\vecskew}[1]{\left[#1\right]_{\times}}
\newcommand{\dotprod}[2]{\left\langle#1,#2\right\rangle}

\def\censorloopword#1 #2\nil{%
  \phantom{#1} 
  \ifx&#2&
    \let\next\relax
  \else
    \def\next{\censorloopword#2\nil}
  \fi
  \next\ignorespaces}

\definecolor{cvprblue}{rgb}{0.21,0.49,0.74}
\usepackage[pagebackref,breaklinks,colorlinks,citecolor=cvprblue]{hyperref}

\def\paperID{359} 
\def\confName{3DV\xspace}
\def\confYear{2025\xspace}

\title{In Flight Boresight Rectification for Lightweight Airborne Pushbroom Imaging Spectrometry}

\author{Julien Yuuki Burkhard\\
{\tt\small julien.burkhard@epfl.ch}
\and
Jesse Ray Murray Lahaye\\
{\tt\small jesse.lahaye@epfl.ch}
\and
Laurent Valentin Jospin\\
{\tt\small laurent.jospin@epfl.ch}
\and
Jan Skaloud\\
{\tt\small jan.skaloud@epfl.ch} \\
École Polytechnique Fédérale de Lausanne, 1015 Lausanne Switzerland\\
}

\begin{document}
\maketitle

\begin{abstract}
    Hyperspectral cameras have recently been miniaturized for operation on lightweight airborne platforms such as UAV or small aircraft. Unlike frame cameras (RGB or Multispectral), many hyperspectral sensors use a linear array or 'push-broom' scanning design. This design presents significant challenges for image rectification and the calibration of the intrinsic and extrinsic camera parameters. Typically, methods employed to address such tasks rely on a precise GPS/INS estimate of the airborne platform trajectory and a detailed terrain model. However, inaccuracies in the trajectory or surface model information can introduce systematic errors and complicate geometric modeling which ultimately degrade the quality of the rectification. To overcome these challenges, we propose a method for tie point extraction and camera calibration for 'push-broom' hyperspectral sensors using only the raw spectral imagery and raw, possibly low quality, GPS/INS trajectory. We demonstrate that our approach allows for the automatic calibration of airborne systems with hyperspectral cameras, outperforms other state-of-the-art automatic rectification methods and reaches an accuracy on par with manual calibration methods.
\end{abstract}

\section{Introduction}

\begin{figure}[t]
    \centering
    \includegraphics[width=\linewidth]{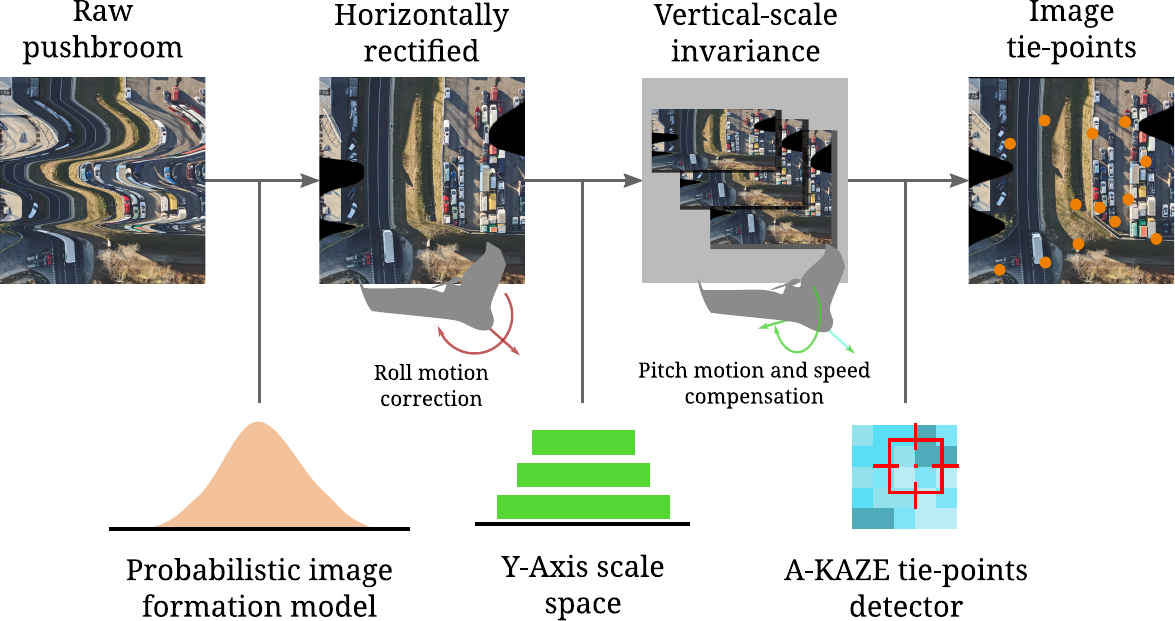}
    \caption{Overview of our proposed process for tie-point extraction in push-broom hyperspectral imagery}
    \label{fig:graphical-abstract}
\end{figure}

Unlike frame cameras, including RGB and multispectral cameras, hyperspectral cameras have not, until relatively recently, been available as payload for light airborne platforms (e.g. UAVs) \cite{NEX2022215}. Lightweight hyperspectral sensors offer many advantages compared to larger, aircraft based sensors, in terms of spatial resolution of the imagery, fast deployment and cost, rendering them very useful for a vast array of applications in the environmental sciences, including but not limited to snow-cover characterization \cite{9686730}, crop and forest monitoring \cite{ASNER1998234} and more. A major issue though, is that lightweight airborne hyperspectral sensors needs to operate a push-broom cameras to maximize spatial and spectral resolution \cite{NEX2022215}. This makes the geo-referencing of the data more challenging, as the camera becomes more reliant on the GPS/INS solution. Orientation determination with inertial measurements especially remain a challenging task for lower end GPS/INS sensors \cite{10591587}. There are two major types of distortions in the output of push-broom sensors which lead to the aforementioned geo-referencing inaccuracy \cite{fang2017two}: high frequency perturbations caused by the platform instability and low frequency drift caused by GPS/INS imprecision. In general, roll motion causes horizontal shift between successive lines in the image, while pitch and speed variation cause deformation along the axis of motion of the camera (see Figure~\ref{fig:graphical-abstract}). 

Imprecision in the geo-referencing of the data, in turn, becomes an issue when multiple data modalities need to be considered together, e.g., Lidar point clouds and hyperspectral images for individual tree species identification \cite{LIU2017170}, or when time series of the data needs to be considered, like for snow cover monitoring \cite{snow_sensing_Dozier}. Another issue with airborne platforms, is that the boresight, i.e., the orientation offset between the projective camera center and the IMU center, needs to be re-calibrated on a regular basis if the GPS/INS receptor is integrated in the airborne platform rather than into the hyperspectral sensor. In fact, most hyperspectral cameras integrate an internal GPS/INS sensor to solve this issue, but this, in turn, makes the payload heavier, requires larger, more expensive drones and limits the capability of the camera to be operated jointly with other payloads. In addition, this often only reduces the amplitude of the boresight, but precise calibration of the sensor is still left to the user \cite{Sankararao2020}.

These shortcomings can be somewhat mitigated using computer vision to rectify the image orientation using bundle adjustment \cite{isprs-archives-XLI-B3-813-2016}, but the processing methodologies for push-broom imagery remain much more complex than the equivalent process for frame imagery now widely available in commercial software suites such as Agisoft Metashape \cite{Metashape} or Pix4D \cite{Pix4D}. There are two main reasons for this additional complexity. First, the motion of the airborne platform can create artifacts in push-broom images, which complicate the automatic extraction of accurate tie points from the data. Second, in push-broom imagery, each line has a different pose, which dramatically increases the number of parameters in traditional bundle adjustment, and introduces certain limits on the geometric constraints that can be used when filtering the data (e.g., epipolar constraints \cite{zhang2021epipolar}), these limitations prevent the use of algorithms such as the 8-point algorithm as is commonly used for frame images, to constrain push-broom imagery.

In this work, we contribute a new method for automatic tie point extraction from hyperspectral images without additional data (i.e., no terrain model and unreliable or imprecise GPS/INS at best):

\begin{enumerate}
    \item We propose to use a probabilistic model of push-broom image formation to rectify the high frequency distortion caused, mainly, by roll motion in the image data. This approach yields sub-pixel accurate estimates of the shift between image lines, and does so without access to a terrain model or GPS/INS data. The conducted experiments demonstrate that our method outperforms the state-of-the-art approaches for roll rectification. As a side benefit, our method allows for the correction of the residual time offset between a GPS/INS trajectory and the hyperspectral data which may exist for low-cost systems without hardware time synchronization.
    \item We propose a y-scale invariant matching-scheme to tackle the distortions caused by pitch motions and speed differences when matching push-broom images. We show that the proposed scheme significantly increases the number of matches and the proportion of inlier matches.
    \item We then show a practical use of the proposed methods by evaluating the quality of boresight calibration of hyperspectral cameras using the proposed visual matches. Tie points generated with our algorithm, coupled with a robust algorithm for boresight estimation, provide an orientation accuracy of around 0.2°, compared to 0.3° for state-of-the-art tie point generation methods.
\end{enumerate}

\section{Prior works}

Processing pipelines for push-broom hyperspectral imagery usually rely only on the GPS/INS solution to georectify image lines from the camera \cite{Sankararao2020, NEX2022215}. To limit the impact of high frequency vibrations, stabilisation gimbals can be used \cite{isprs-archives-XLII-2-W6-379-2017} but they pose other issues (i.e., time varying boresight and lever arm between the GPS/INS receptor and the camera), most notably when the GPS/INS receptor is placed on the airborne platform.

Historically, strong straight features (e.g. roads) had to be manually identified to correct for distortion caused by aircraft motion (mostly roll motion) \cite{jensen2008single}, but such methods are not feasible in the absence of a straight, linear object in the image, and are complex to automatize. In addition to linear features, Ground Control Points (GCPs), marked with recognizable patterns that can be identified in the images and for which spatial coordinates are measured in the terrain, can also be used as constraints \cite{doi:10.2747/1548-1603.48.3.416}. Alternatively to GCPs, which require an intervention in the terrain, image correspondences or tie points can be extracted from overlapping image regions. Tie points have been widely used for hyperspectral sensor calibration, either extracted from overlapping hyperspectral cubes \cite{8630822}, or between hyperspectral cubes and frame images \cite{isprs-archives-XLI-B3-813-2016}. However, the reliable extraction of tie points can be challenging in the presence of distortion in hyperspectral imagery, which is a bit of a conundrum when such tie points are required to rectify  such deformations in the first place. In such cases, the image can be pre-rectified using only the raw solution from the GPS/INS sensor, but without prior boresight calibration, when the GPS solution is noisy or when a time offset exist between the cameras and the GPS/INS sensor, artifacts can be generated in the image during the rectification process \cite{8326499}.

To address this, it has been shown that the pixel cross-correlation between successive lines can be used to correct the horizontal displacement caused by the aircraft roll motion \cite{fang2017two}. This makes it possible to extract tie-points from the partially rectified hyperspectral cubes, without requiring any form of direct geo-referencing of the data and thus without knowing the camera boresight. We propose to use a similar approach for horizontal displacement correction, but we replace the cross-correlation with a Bayesian probabilistic model, described in Section~\ref{sec:meth:horizshift}. Our model provides multiple benefits over the state of the art. For example, it predicts the horizontal shift between successive image scan lines with sub-pixel accuracy, due to the improved modelling of the stochastic properties of the image and includes a prior acting as a regularizer for the solution.

For tie point extraction in hypespectral imagery, specific variants of popular descriptors such as SIFT have been proposed \cite{yu_spatial-spectral_2021}, but in practice, it has been observed that traditional descriptors yield good results without the need to be adapted \cite{8630822}. In this work, we used the off the shelf A-KAZE matching pipeline \cite{alcantarilla2011fast} to detect tie points after rectifying the images.

\section{Methodologies}
\label{sec:meth}

The proposed method aims to obtain tie points for hyperspectral push-broom imagery in the presence of either an unstable platform, inaccurate GPS/INS measurements, time tagging offset w.r.t GPS time and/or unknown boresight between the GPS/INS system and the camera. We split the main problem into three sub-problems: 1) correcting of the horizontal shift caused by roll, 2) removing the ambiguity in vertical scaling due to pitch motion and speed and finally 3) tie point extraction. 

Section~\ref{sec:meth:datasets} presents the datasets used in our experiments. Section~\ref{sec:meth:horizshift} presents our method for horizontal shift correction, based on a probabilistic model. Section~\ref{sec:meth:vertscale} presents our approach for dealing with the ambiguity in vertical scaling of the raw data. Finally, Section~\ref{sec:meth:tiepoints} presents our approach for tie-point extraction.

\subsection{Datasets}
\label{sec:meth:datasets}

\begin{figure} [b]
    \centering
    \begin{subfigure}[t]{0.22\textwidth}
        \centering
        \includegraphics[height=70pt]{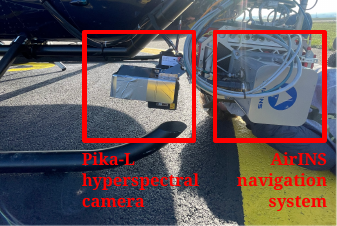}
        \caption{Camera and navigation system}
    \end{subfigure}
    \hfill
    \begin{subfigure}[t]{0.22\textwidth}
        \centering
        \includegraphics[height=70pt]{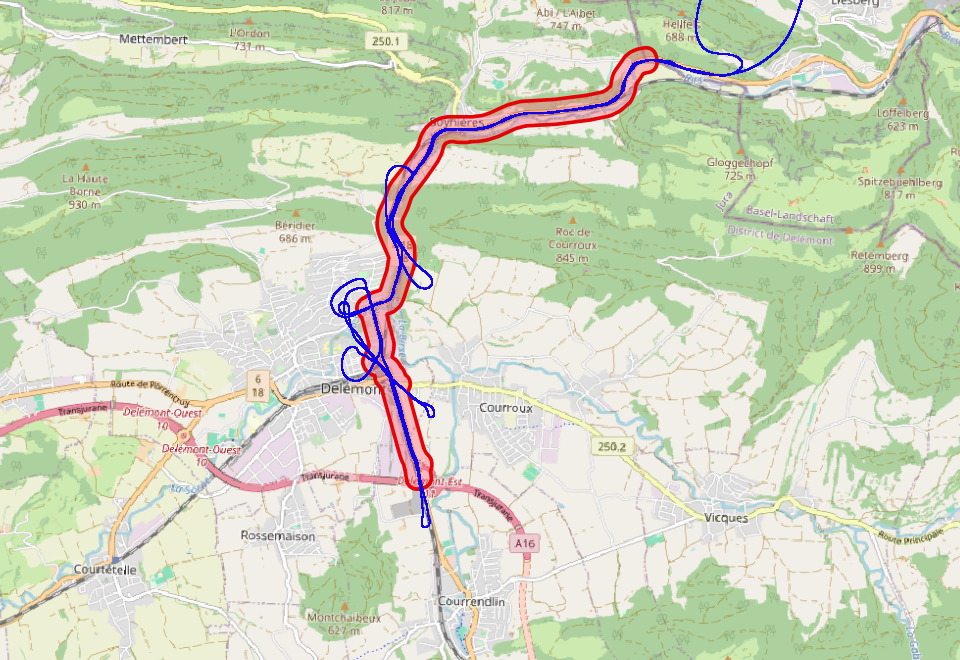}
        \caption{Flight path above Delemont with area of interest and flight path}
    \end{subfigure}
    \caption{Dataset acquisition in Delemont (CH), map background \textcopyright~OpenStreetMap contributors}
    \label{fig:dataset_acquisition}
\end{figure}

The main dataset for this work was acquired in the region of Delemont, in Switzerland, see Figure~\ref{fig:dataset_acquisition}. We used a Resonon Pika-L portable hyperspectral camera, mounted on a helicopter, alongside a reference navigation grade GPS/INS system with a Javad GNSS receiver coupled with an iXblue AirINS inertial platform. A lidar and RGB camera were also used during the flight, but we will not use these data in this work. The Pika-L comes bundled with a GPS-INS system, the SBG Ellipse-N, which is representative of the quality one could expect by flying the camera on a UAV, while the output of the AirINS can be used as a ground truth reference. The Pika-L has a field of view of 36.5° and each line contains 900 spatial pixels. The nominal focal length is 1345 pixels.

We used a hardware output pulse triggered for each scan line exposed by the Pika-L to time-tag them with the absolute GPS time of the Javad receiver. The trajectory produced by the AirINS was also synchronized with respect to absolute GPS time using the same GNSS receiver, while the Ellipse-N trajectory is given with respect to the system time of the Pika-L camera. An issue that arose, is that the Pika-L sent more pulses than lines were saved, probably due to hardware limitations of the camera. Nonetheless, since our method for horizontal shift rectification, presented in Section~\ref{sec:meth:horizshift}, does not require GPS/INS data, but the horizontal shift is correlated with roll, we were able to recover the correct time offsets between the system time of the camera and absolute GPS-time. We provide more details as to the methodology applied to do so in the supplementary material. In addition, due to small delays in the onboard computer, we observed small time offsets between the Ellipse-N trajectory and the hyperspectral data. We used the same methodology to correct for these offsets as well.

As a check, we also conducted experiments with a publicly available UAV based, hyperspectral dataset \cite{Kim_Alaska_UAV_Hyperspectral} to provide additional evaluation for our horizontal shift correction model described in Section~\ref{sec:meth:horizshift}. However,this dataset is limited to straight, parallel flight lines, thus we could not use it to evaluate our boresight calibration method, and rely only on our dataset for the evaluation of the tie point extraction and calibration methods described in Sections~\ref{sec:meth:vertscale}, \ref{sec:meth:tiepoints} and \ref{sec:meth:boresight}.

\subsection{Horizontal shifts and probabilistic image formation models}
\label{sec:meth:horizshift}

At the core of our approach, we consider that the value of two successive lines in the push-broom images, represented here as a vector $\vec{I}$, follow approximately a normal distribution:

\begin{equation}
    \label{eq:multivariatenorm}
    \vec{I} \sim \mathcal{N}\left(\vec{\mu}, \matr{\Sigma}\right).
\end{equation}

\noindent This derives from the assumption that the image generation can be approximated by a Gaussian Process, an assumption which is not perfect, but has been shown to be sufficiently accurate in practice when correct covariance functions are used \cite{5995713}. If we have a prior on how different pixels are correlated based on the relative poses of the camera between the two lines, the observing $\vec{I}$ yields insight about the relative orientation of the camera between two lines by inverting the conditional probability distribution between the pose and the pixel values. A straightforward way of doing so is by using the Bayesian formalism and probabilistic graphical models \cite{Buntine_1994}.

\begin{figure}
    \centering
    \includegraphics[width=0.5\linewidth]{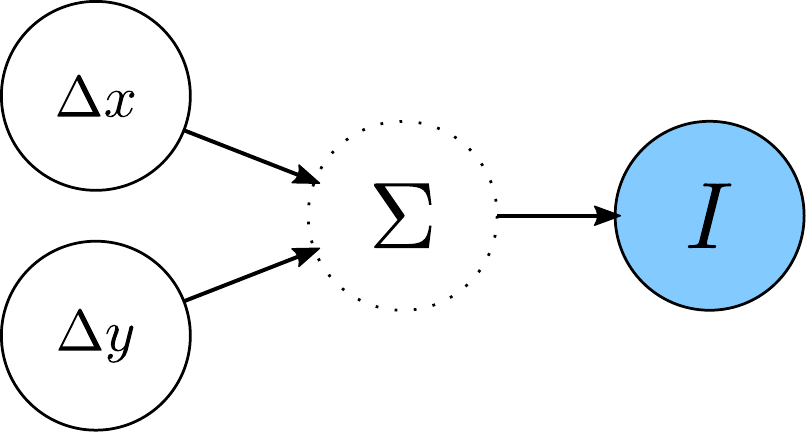}
    \caption{Probabilistic graphical model, as defined in \cite{Buntine_1994}, used for the estimation of the horizontal shift. Blue circles are observed variables, white circles latent variables and dashed circles deterministic functions}
    \label{fig:pgm_estimate}
\end{figure}

To simplify the interpretation of the model output, we do not consider the whole 6-DOF pose in 3D, but instead only consider the resulting horizontal (cross-path) shift $\Delta x$ and vertical (along-path) shift $\Delta y$. For a nadir facing camera and flat terrain, the number of pixels per radian as a function of the pixel position is given by:

\begin{equation}
    \dfrac{f}{cos^2\left( arctan\left(\dfrac{x}{f}\right) \right)},
\end{equation}

\noindent with $f$ being the focal length in pixels and $x$ is a given pixel coordinate. This means that the relative shift at a given pixel position with respect to a shift at the optical center of the camera is given by:

\begin{equation}
    \dfrac{1}{cos^2\left( arctan\left(\dfrac{x}{f}\right) \right) } - 1,
\end{equation}

\noindent which translates to a relative error of 10.9\% on the edge of the image for the Pika-L, and an average error per line of 3.6\%. This is not fully negligible but still acceptable knowing the targeted estimation error is 0.1 pixels, or 10\%.

We build our graphical probabilistic model as shown in Figure~\ref{fig:pgm_estimate}. We set the prior to be $\Delta x \sim \mathcal{N}(0, 0.5)$ and $\Delta y \sim \text{Exp}(1)$. The values of the standard deviation and rate were chosen empirically from the raw IMU data. 

The likelihood model is given, following Equation~\ref{eq:multivariatenorm}, by:

\begin{equation}
    p(\vec{I} | \Delta x, \Delta y) = \mathcal{N}\left(\vec{\mu}, \matr{\Sigma}(\Delta x, \Delta y)\right),
\end{equation}

\noindent with $\vec{\mu}$ the mean of the pixel value intensity, and the function $\matr{\Sigma}(\Delta x, \Delta y)$ computed using a Matérn covariance kernel \cite{10.5555/944790.944815}. With this model, the covariance between two pixels separated by a horizontal distance $d$ is given by:

\begin{equation}
    \sigma^2\left(1+\frac{\sqrt{3}\sqrt{(\Delta x + d)^2 + \Delta y^2}}{l}\right) e^{\left(-\frac{\sqrt{3}\sqrt{(\Delta x + d)^2 + \Delta y^2}}{l}\right)},
\end{equation}

with $\sigma^2$ the variance of the image pixel intensity and $l$ a length factor that can be calibrated by evaluating the covariance of pixels on the same row (thus without distortion). We set $\Delta x = \Delta y = 0$ for pixels on the same row. The Matérn covariance function ensures that the covariance between two pixels decrease effectively with the distance, while ensuring that $\matr{\Sigma}(\Delta x, \Delta y)$ stays positive definite for all values of $\Delta x$ and $\Delta y$, which an arbitrary covariance function cannot guarantee.

In the end, we select the estimator for $\Delta x$ and $\Delta y$ to be:

\begin{equation}
     \Delta \hat{x}, \Delta \hat{y} = \argmax_{\Delta x,\Delta y} p(\vec{I}|\Delta x,\Delta y)p(\Delta x)p(\Delta y).
 \end{equation}

\noindent A closed form expression for $log\left( p(\vec{I}|\Delta x,\Delta y)p(\Delta x)p(\Delta y) \right)$ and its derivative exist, but solving for the minima is not practical without a numerical solver. Thus, to compute $\Delta \hat{x}$ and $\Delta \hat{y}$, we rely on a numerical solver. We give more details in the supplementary material on the solving method we use.

A consequence of this approach, is that the asymptotic time complexity of our method, $\mathcal{O}(n^3)$, is slightly worse than basic correlation $\mathcal{O}(s \cdot n)$. Here, $s$ is the number of shifts under consideration and $n$ the number of pixels in line patches considered at once. We assume $s < n$ when all pixels in the line are considered at once. A mitigation strategy is instead to consider multiple patches of fixed size and estimate $\Delta \hat{x}$ and $\Delta \hat{y}$ by assuming these patches are independent. Beside reducing the time complexity of our method to $\mathcal{O}(n \cdot p^2)$ (with $p$ the patches size), this approach also ensures that a strong feature in the image cannot bias the estimator too much. Nevertheless, when the size of the patches is reduced, the capacity of our method to detect larger shifts with precision decreases, so there is a trade-off that must be calibrated. In this work we used individual patches of $16px$ which was a value identified experimentally the details of which we provide in the supplementary material. 

\subsection{Vertical scale invariance}
\label{sec:meth:vertscale}

Considering the observed fact that $\Delta \hat{x}$ yields quite a good estimate of the horizontal shift and is very correlated with the roll motion of the airborne platform, $\Delta \hat{y}$ is usually less predictable and yields very little information about the local distortion along the travel direction. This is because when transitions in the landscape occur along the axis of travel, the local correlation hypothesis is no longer valid. In that case, $\Delta \hat{x}$ is only marginally affected while, due to the nature of the push-broom sensor, $\Delta \hat{y}$ is strongly affected. For our intended application, we need a different method to deal with the vertical distortion in the hyperspectral data.

\begin{figure}[t]

    \begin{subfigure}[t]{0.48\columnwidth}
        \centering
        \includegraphics[width=\textwidth]{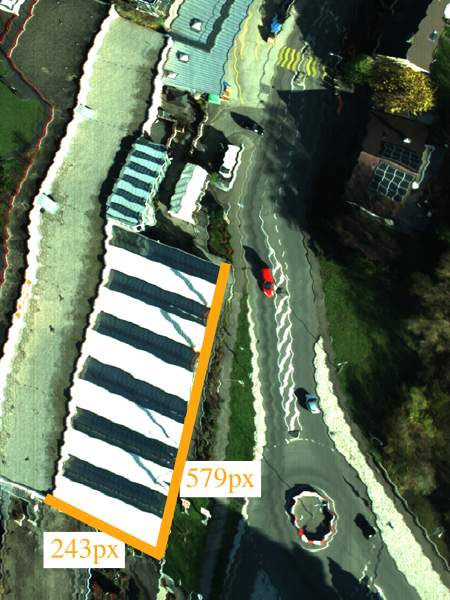}
        \subcaption{Parallel crossing}
        \label{fig:y-scaling-effect:horz}
    \end{subfigure}
    \hfill
    \begin{subfigure}[t]{0.48\columnwidth}
        \centering
        \includegraphics[width=\textwidth]{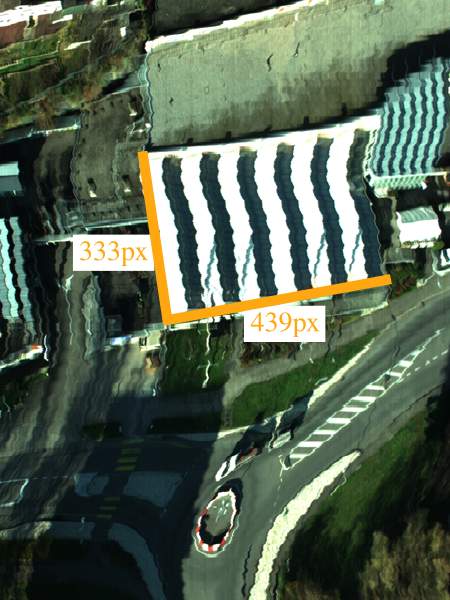}
        \subcaption{Perpendicular crossing}
        \label{fig:y-scaling-effect:perp}
    \end{subfigure}

    \caption{The effect of y-scaling, where the aspect ratio of the image changes when the airborne platform flies parallel (a) or perpendicular (b) to the road.}
    \label{fig:y-scaling-effect}
\end{figure}

This is because, for crossings flight lines, i.e., when tie points are the most useful for calibration purposes, the variable speed of the aircraft can lead to distortion in the push-broom pixels along the direction of travel, see Figure~\ref{fig:y-scaling-effect}. These distortions, in turn, complicate the process of matching keypoints as they act in opposite directions in the two crossing flight lines.

To deal with possible distortions in the $y$ direction, we modified the A-KAZE scale invariance feature pyramid, such that the scale in the $x$ and $y$ directions are now explored independently in the scale pyramid. This anisotropic scale pyramid trades speed and complexity of the algorithm for invariance against $y$ scaling. Nonetheless, in aerial imaging, the aircraft usually flies at constant height above the ground, so very few levels in the $x$ direction need to be considered, generally a single level and hardly ever more than 2 different levels. In our study we only considered a single scale level in the $x$ direction, and the whole feature pyramid in meant to deal with the $y$ scaling of the data.

\subsection{Tie point extraction and filtering}
\label{sec:meth:tiepoints}

Once the image is rectified as described in Section~\ref{sec:meth:horizshift}, it becomes possible to use a traditional pipeline to extract matching tie points. We used the A-KAZE method \cite{alcantarilla2011fast}, which is known to yield fast and accurate results \cite{HSI_KAZE}, and has the benefits to be open source. Initial testing on our data hinted that A-KAZE is a better alternative as it yields more matches and a similar proportion of inliers than SIFT. We also compared with ORB \cite{6126544}, which is often the recommended matching method \cite{8346440}. While ORB yields more matches, a much higher proportion of them are outliers, which complicates downstream tasks.

Once the tie points matches are produced, a significant portion of them can still be incorrect (i.e. the algorithm matches two pixels representing different elements in the scene). In the case of frame images, geometric constraints and the RANSAC algorithm \cite{lacey2000evaluation}, chief amongst them being the epipolar constraints, can be used to detect and remove outliers. In the case of un-calibrated pushbroom camera, assuming the GPS/INS measurement is either unreliable or imprecise, this is not feasible. Instead, we consider two rectified chunks of the pushbroom hyperspectral data and then assume that a generic homography exists between the points in chunk 1 and chunk 2. In other words, we assume the existence of a unique matrix $\matr{H}$ such that:

\begin{equation}
    \label{eq:homography}
    \vec{pt}_{1,i} ~\propto~ \matr{H} \vec{pt}_{2,i},
\end{equation}

\noindent where $\vec{pt}_{1,i}$ and $\vec{pt}_{2,i}$ are matching points in chunks 1, respectively 2, in homogeneous coordinates. The assumption here is that there exists a perspective transformation between the sets of inlier points in the two respective chunks. 

The coefficients of $\matr{H}$ can be estimated with at least 4 points using Equation~\ref{eq:homography} and the direct linear transform algorithm \cite{abdel2015direct}. Based on this model, a RANSAC loop can be built. In the RANSAC loop, we consider a point to be an inlier if it lies at a distance of at most 60 pixels, which is very tolerant to account for the fact that our homography model is only an approximation.

\subsection{Hyperspectral camera boresight calibration}
\label{sec:meth:boresight}

The main parameter we want to calibrate for a hyperspectral camera is the boresight. Intrinsic parameters like the focal length and lens distortion can be calibrated in advance, meaning that we have a strong prior for them. The lever arm between a GPS/INS sensor and the camera can be measured on the platform before the flight with good accuracy (relative to the scale of the scene). On the other hand, the boresight, which is key for precise geo-referencing of the mapping data, cannot be measured. Tie points offer a good system for the boresight calibration from the mapping data itself, limiting the need for a separate calibration procedure before flight. Due to geometric constraints, it still imposes requirements on the flight path, as crossing lines are required to disambiguate the projected effects between the different axes of rotation.

The tie points can also be used for the full co-optimization of the trajectory and camera poses with full mounting offsets. This is usually done by including visual constraints in a Kalman filter \cite{isprs-archives-XLI-B3-813-2016} or Factor Graph Optimization \cite{Jospin2023-11-18}. But this relies on a more complex optimization model and is out of the scope of this work.

To estimate the boresight, we minimize the residuals of epipolar constraints for each pair of tie points. We parameterized the boresight as an axis angle $\vec{r}$, in which the direction indicates the axis of rotation, and the norm indicates the angle. This is in fact an element of $\liealgebra{so}(3)$, the lie algebra associated with the special orthogonal group $\liegroup{SO}(3)$. This means that the corresponding rotation matrix $\matr{R}$ is given by the matrix exponential $e^{\vecskew{\vec{r}}}$, with $\vecskew{\vec{r}}$ the matrix representing the cross product with $\vec{r}$. The epipolar constraint for a pair $i$ is written as:

\begin{equation}
\label{eq:epipolar_constraint}
    \dotprod{e^{\vecskew{\vec{r}}} \vec{v}_{1,i} \times \matr{R}_i e^{\vecskew{\vec{r}}} \vec{v}_{2,i}}{\vec{t}_i} = 0,
\end{equation}

\noindent with $\vec{v}_{1,i}$ and $\vec{v}_{2,i}$ being the ray direction in the camera frame at pose 1, respectively 2, $\matr{R}_i$ being the relative rotation from pose 2 to pose 1 and $\vec{t_i}$ being the translation between pose 1 and 2.  $\matr{R}^1_2$ and $\vec{t}$ are taken from the GPS/INS output, and as such would be noisy, but the estimate of $\vec{r}$ is obtained with a very large number $n$ hundred, if not thousands tie points, lowering the noise level with a factor $\mathcal{O}(\sqrt{n})$ as long as no bias is present.

Equations~\ref{eq:epipolar_constraint} can be solved using a numerical solver like Gauss-Newton or Levenberg–Marquardt. We used Gauss-Newton, as the problem proved to be well behaved and converges to the solution without a need for additional regularization.

The Gauss-Newton algorithm solves an over-determined set of equations in the least square sense, meaning that a few outliers can have an out sized effect on the solution. To limit the impact of outliers, it can be useful to wrap the residuals of the equations in a non-linear kernel. A popular option is the Huber Loss, which is quadratic close to zero and linear further away:

\begin{equation}
    H_\delta (x) = \begin{cases}
 \frac{1}{2}{x^2}                   & \text{for } |x| \le \delta, \\
 \delta \cdot \left(|x| - \frac{1}{2}\delta\right), & \text{otherwise.}
\end{cases},
\end{equation}

\noindent with $\delta$ an hyper-parameter of the loss. We selected $\delta = 1/4$. The updated robust constraints are:

\begin{equation}
\label{eq:robust_epipolar_constraint}
    \sqrt{H_{1/4} \left( \dotprod{e^{\vecskew{\vec{r}}} \vec{v}_1 \times \matr{R}^1_2 e^{\vecskew{\vec{r}}} \vec{v}_2}{\vec{t}} \right)} = 0,
\end{equation}

\noindent with the square root to account for the fact that the Gauss-Newton algorithm will minimize the square of our constraints.

\section{Experiments and evaluation}
\label{sec:results}

In this section, we evaluate the different methods that were described in Section~\ref{sec:meth}. First, we evaluate our proposed probabilistic model for horizontal shift correction in Section~\ref{sec:results:horizontal}. Then, in Section~\ref{sec:results:tiepoints_and_filtering}, we evaluate how the proposed horizontal shift correction and y-scale invariant approach impact the production of tie points, and the proportion of outliers filtered out. Finally, in Section~\ref{sec:results:boresight}, we measure the quality of boresight calibration as a function of the produced automated correspondences.

\subsection{Horizontal shift correction}
\label{sec:results:horizontal}

To evaluate the quality of the proposed horizontal-shift correction method, we compare its performance against the state of the art method by Fang \etal \cite{fang2017two}. In the main paper we present the results on our main dataset. Results on the additional dataset \cite{Kim_Alaska_UAV_Hyperspectral} are reported in the supplementary material.

\begin{figure}
    \centering
    \includegraphics[width=\linewidth]{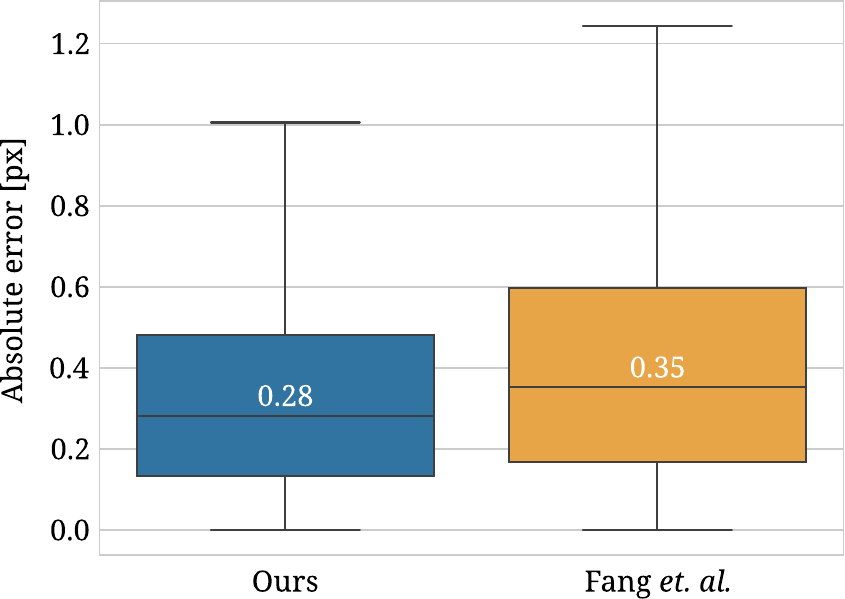}
    \caption{Horizontal shift error distribution (according to to $\Delta x$ predicted by GPS/INS)}
    \label{fig:horizontal-shift-error-distribution}
\end{figure}

\begin{figure*}

    \begin{subfigure}[t]{0.48\textwidth}
        \centering
        \includegraphics[width=\textwidth]{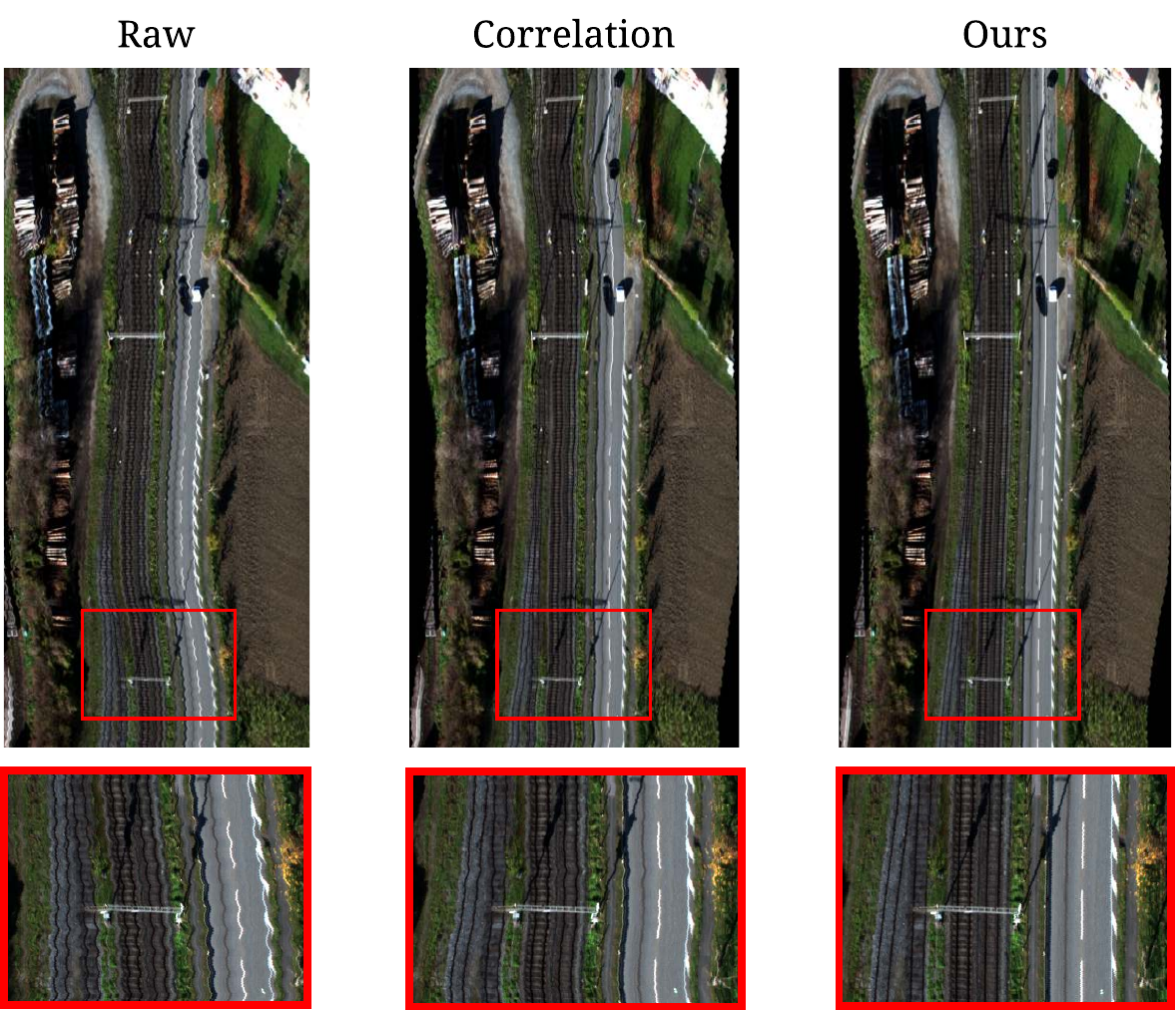}
        \subcaption{Rectification quality: less deformations are visible along the road and railway with our method.}
        \label{fig:visual_results_horizontal_shift:qual}
    \end{subfigure}
    \hfill
    \begin{subfigure}[t]{0.48\textwidth}
        \centering
        \includegraphics[width=\textwidth]{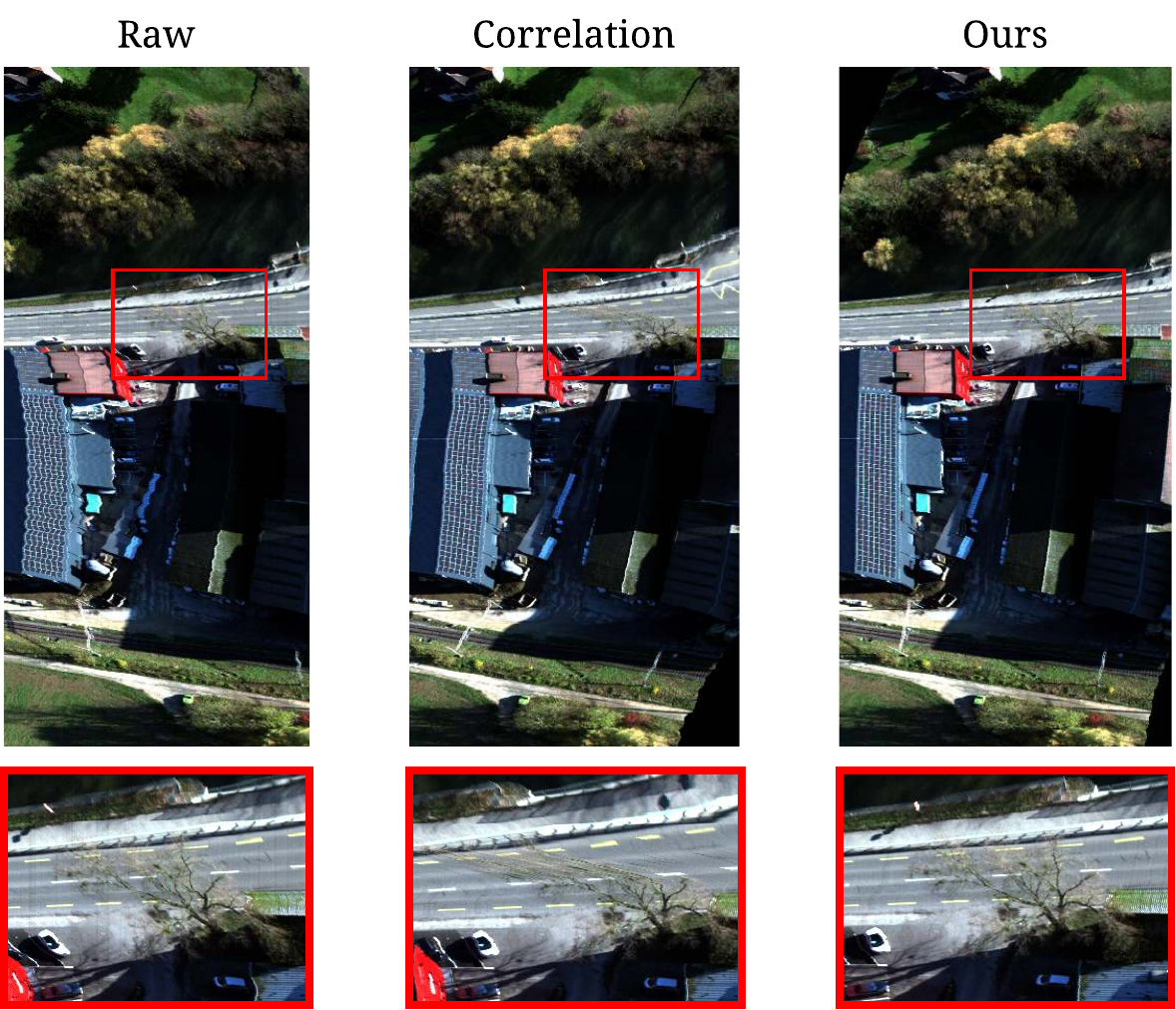}
        \subcaption{Resistance to local attractors: features on the road seem to distort the tree in the correlation based rectification, not with our method.}
        \label{fig:visual_results_horizontal_shift:res}
    \end{subfigure}

    \caption{Our method is both more accurate than the state of the art (a), but also, the Bayesian prior ensure that the solution is not too strongly impacted by local attractors (b)}
    \label{fig:visual_results_horizontal_shift}
\end{figure*}

We use the AirINS GPS/INS reference solution to estimate the ground truth horizontal displacement $\check{x}$. We compute the shift in pixels of the features on the ground, based on the digital surface model of the terrain. We then average that shift to get $\check{x}$. Note that only the regions within the mapping area were included, regions outside of the mapping areas highlighted in Figure~\ref{fig:dataset_acquisition}, where the helicopter is turning, are excluded.

Overall, for the Delemont dataset, our method reaches an average accuracy, measured by the RMSE, of 0.85 pixels. Note that the RMSE is very strongly influenced by a small set of outliers. When the median is considered instead, the accuracy is 0.28 pixels.

For comparison, the previous state of the art method by Fang \etal \cite{fang2017two}, based on the cross correlation of successive lines, reach a RMSE 0.93 pixels and a median accuracy of 0.35 pixels.

Figure~\ref{fig:horizontal-shift-error-distribution} shows the distribution of error with the median, as well as the 5\%, 25\%, 75\% and 95\% quantiles.

Observations in the rectified images (Figure~\ref{fig:visual_results_horizontal_shift}) show there are two main reasons why our method reaches this accuracy:

First, since our model is continuous (able to detect sub-pixel shifts), and based on a better approximation of the image formation model, it reaches a better rectification quality. This is especially visible in Figure~\ref{fig:visual_results_horizontal_shift}a, where both the road and railway are still distorted when correlation is used, but close to no deformation is visible with our approach.

Second, as our approach is based on the Bayesian formalism, it includes a prior for the latent variable. This prior in turn has a regularizing effect \cite{9756596}, which prevents the solution from collapsing to a slightly more optimal but yet less likely shift. This effect is especially visible in Figure~\ref{fig:visual_results_horizontal_shift}b, where the texture on the road across the image seems to have just the right tilt to cause strong and visible deformations on the other elements of the image, here a tree. With our method, no such distortion is apparent.

\begin{table*}[t]
    \centering
    \begin{adjustbox}{width=\textwidth,center}
\begin{tabular}{llrrrr}
\hline
\textbf{Method}                  & \textbf{Y-scale invariant}                & \textbf{Keypoints}             & \textbf{Matches}              & \textbf{Inlier matches}       & \textbf{Selected Homography Ransac}         \\ \hline
{\color[HTML]{9B9B9B} Reference} & {\color[HTML]{9B9B9B} -} & {\color[HTML]{9B9B9B} -} & {\color[HTML]{9B9B9B} (manual) 949} & {\color[HTML]{9B9B9B} 100.0\%} & {\color[HTML]{9B9B9B} -} \\
Raw                              & No                                  & 288'024                        & 2'213                         & 16.0\%                          & 871 (acc. 69.9\%)                           \\
Fang \etal \cite{fang2017two}                   & No                                  & 310'433                        & 3'088                         & 12.5\%                        & 1'145 (acc. 69.6\%)                         \\
Fang \etal \cite{fang2017two}                     & Yes                                 & 500'971                        & 12'691                        & 58.8\%                        & 7'920 (acc. 92.2\%)                         \\
Ours                             & No                                  & 309'187                        & 2'380                         & 15.0\%                          & 917 (acc. 72.7\%)                           \\
Ours                             & Yes                                 & 499'947                        & 14'459                        & \textbf{69.2\%}               & \textbf{10'289 (acc. 94.0\%)}               \\ \hline
\end{tabular}
    \end{adjustbox}
    \caption{Number of tie points and inliers for different methods. Bold is best proportion of inliers (excluding reference)}
    \label{tab:tie_points_extraction_results}
\end{table*}

\subsection{Tie points detection and filtering}
\label{sec:results:tiepoints_and_filtering}

Now, the main question is, does the increase in accuracy of the horizontal shift estimate lead to better tie points? To check that, we ran our tie point detector on the raw data, the data rectified using the correlation based approach described by Fang \etal \cite{fang2017two} and the data rectified using our method. We also used the ground truth GPS/INS data to rectify the hyperspectral data and manually selected 949 tie points. These manual tie points will be used later as comparison points when evaluating the accuracy of the boresight calibration.

Once the tie points were obtained by the different methods, we manually categorized them as either inliers, when they are visibly pointing at the same object, or outliers, when they are pointing to different objects. Around 1'000 tie points were labeled in this way for each sets of points. This allows to compare the accuracy of the different methods, in term of proportions of inliers that are present.

Finally, we used the homography based RANSAC filtering approach described in Section~\ref{sec:meth:tiepoints} to filter the outliers again, estimating the proportion of inliers after filtering in SOTA as well as our method.

The results, reported in Table~\ref{tab:tie_points_extraction_results}, show clearly that, 1)  using a y-scaling invariant scheme, like the one we propose, is very beneficial. Both rectification methods saw an increase in the number of tie points and the proportion of inliers when the y-scaling invariant scheme was used. 2) The proposed filtering method based on a homography model yield a significant increase in the proportion of inliers, but is not perfect. In fact, horizontal shift without y-scaling invariance seems to be counterproductive compared to running the detector on the raw data, probably because the horizontal correction re-aligned patterns on the road, which in turn increased the amount of self-similar texture areas where the algorithm could fail. 3) Our method reaches a higher proportion of inliers, both before (\~70\%) and after filtering (\~95\%), showing the benefits of a better horizontal rectification algorithm.

\begin{table*}[t]
    \centering
    \begin{adjustbox}{width=\textwidth,center}
    \begin{tabular}{lllrr}
\hline
\textbf{Method} & \textbf{Filter method} & \textbf{Optimization kernel} & \textbf{Bootstrap standard error} & \textbf{Error with reference} \\ \hline
{\color[HTML]{9B9B9B} Reference  }     & {\color[HTML]{9B9B9B} Manual  }    & {\color[HTML]{9B9B9B} $\ell 2$}  & {\color[HTML]{9B9B9B} 0.17°}  & {\color[HTML]{9B9B9B}  -  } \\
Raw             & Homography Ransac      & $\ell 2$                 & 5.77°                 & 10.35°                        \\
Raw             & Homography Ransac      & Huber               & 1.33°                 & 5.72°                         \\
Fang \etal \cite{fang2017two}      & Homography Ransac      & $\ell 2$                  & 4.92°                 & 20.97°                        \\
Fang \etal \cite{fang2017two}      & Homography Ransac      & Huber               & 0.30°                 & 0.34°                         \\
Ours            & Homography Ransac      & $\ell 2$                  & 3.59°                 & 11.37°                        \\
Ours            & Homography Ransac      & Huber               & \textbf{0.22°}        & \textbf{0.12°}                \\ \hline
\end{tabular}
    \end{adjustbox}
    \caption{Comparisons of different approach for boresight calibration, with expected angular error and error compared to reference. Bold is best accuracy (excluding reference)}
    \label{tab:boresight_calibration_results}
\end{table*}

Overall, our method brings a higher proportion of inlier tie points. In the next section, we investigate if this higher proportion of inlier tie points translates to a better estimation of the boresight calibration, which is the end application we are interested in.

\subsection{Camera boresight calibration}
\label{sec:results:boresight}

To estimate the uncertainty associated with the obtained set of tie points, we use bootstrapping \cite{Efron1985}, i.e., estimate the uncertainty associated with an input distribution by using resampling in the set of measurements. While Laplace's Approximation \cite{mackay2003information} is often used to get the uncertainty of an estimate obtained via Gauss-Newton's optimization (the covariance of the estimated posterior is just the inverse of the Hessian of the optimization problem), it also assumes that the errors in the inputs are normally distributed. Laplace's Approximation is also meant to estimate the uncertainty given a set of measurements, not the uncertainty associated with the generation of the measurements themselves. To account for the fact that the error distribution of the obtained tie points is probably heavy tailed and assumed to be arbitrary, we use bootstrapping instead. 

For each set of filtered points shown in Table~\ref{tab:tie_points_extraction_results}, including the 949 reference points, we randomly select 500 and solve the problem of boresight calibration. For each tie point set, except for the reference which we know contains no outliers, we compare the results when using either the $\ell 2$ or the Huber kernel for the Gauss-Newton optimization.

We repeat this experiment 100 times. From the 100 generated samples, we compute the mean, as well as the expected angle deviation from the mean. The angle deviation between two rotations $\vec{r}_1$ and $\vec{r}_2$, with $\vec{r}_1$ and $\vec{r}_2$ expressed as axis angles, can be computed as:

\begin{equation}
\label{eq:angle_deviation}
    arccos \left( \dfrac{Tr \left(e^{-\vecskew{\vec{r}_1}} e^{\vecskew{\vec{r}_2}} \right) - 1 }{2} \right) .
\end{equation}

\noindent We then compare the boresight estimate for each generated tie point set and compare it with the estimate obtained with manual tie points (which we will use as ground truth), computing the deviation with Equation~\ref{eq:angle_deviation} again.

The results, shown in Table~\ref{tab:boresight_calibration_results}, highlight the clear benefits of our method. we get a 30\% reduction in the uncertainty associated with the tie point generation process compared with the method of Fang \etal \cite{fang2017two}, and a 50\% reduction in the error compared to the ground truth (i.e., the reference obtained with manual tie points). The caveat is that the use of a robust regression kernel like the Huber loss, seems to be mandatory, as without, the remaining outliers seems to have an out-sized effect, but with less than 10\% outliers, the robust estimator remains stable.

\section{Conclusion and future work}

In this work, we demonstrated that a method for push-broom hyperspectral horizontal shift rectification method based on probabilistic modelling of the image formation process, coupled with a y-scale invariant tie points extraction algorithm and robust regression allows to solve the boresight calibration problem with an accuracy on par with manually curated tie points. Besides "in the wild" calibration of the boresight for lightweight hyperspectral push-broom camera, our method of tie points generation could be used for other problems like global GPS-INS adjustment or other re-calibration procedures for the camera parameters.

Since our method for tie point production does not require external input, another interesting research direction for future works is multiple sensors co-registration.

\section{Acknowledgements}

Julien Burkhard conducted most of the experiments described in this paper under the supervision of Jesse Lahaye and Laurent Jospin. Experiments were designed by Laurent Jospin. Processing tools for hand labelling of the ground truth data were developed by Laurent Jospin. The overall research was directed by Jan Skaloud. Thanks to Julien Vallet and Sixense Helimap for facilitating the acquisition of our dataset. Thanks to Davide Gucci for their valuable inputs.


{
    \small
    \bibliographystyle{ieeenat_fullname}
    \bibliography{main}
}

\end{document}


\maketitle

\section{Time offset calibration between GPS/INS and camera using horizontal offset estimate}

\begin{figure}
    \centering
    \includegraphics[width=\textwidth]{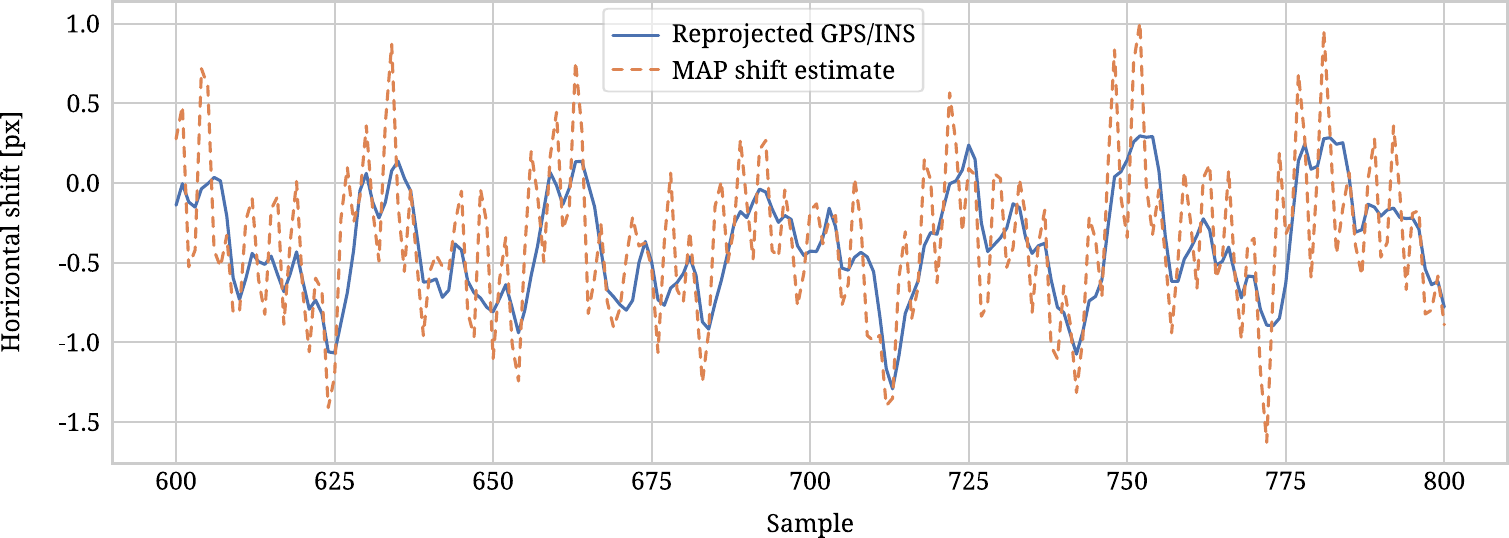}
    \caption{Sychronized signals between our horizontal shift estimate and the roll pixel projection from the AirINS.}
    \label{fig:timing-cross-correlation}
\end{figure}

As presented in the main paper, the horizontal shifts estimates $\Delta x[t]$ we can obtain with our Bayesian model are highly correlated to the roll motion of the airborne platform, see Figure~\ref{fig:timing-cross-correlation}. As such, it is possible to estimate the time offset $\Delta t$ between the sequence of horizontal shifts $\Delta x[t]$ and the roll angular speed $\text{roll}[t]$ time series using simple cross-correlation \cite{blackledge2006digital}:

\begin{equation}
\label{eq:time_correlation}
    \Delta \hat{t} = \argmax_{\Delta t}  \sum_t \Delta_{t} \text{roll}[t] \cdot \Delta x[t+\Delta t].
\end{equation}

\begin{figure}
    \centering

    \begin{subfigure}[t]{0.4\textwidth}
        \centering
        \includegraphics[width=\textwidth]{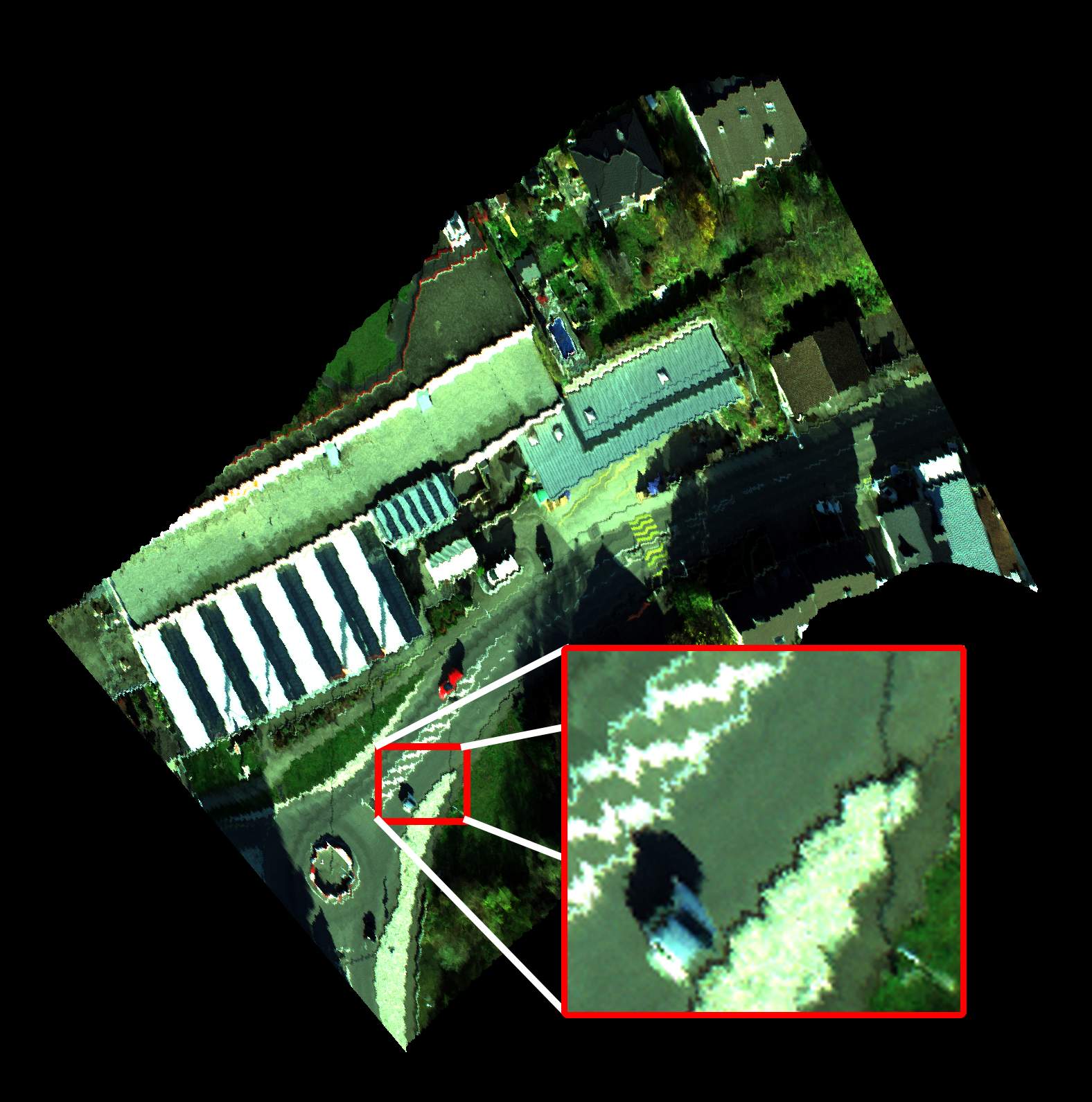}
        \subcaption{Without time synchronization}
        \label{fig:time_sync_results:without}
    \end{subfigure}
    \hspace{20pt}
    \begin{subfigure}[t]{0.4\textwidth}
        \centering
        \includegraphics[width=\textwidth]{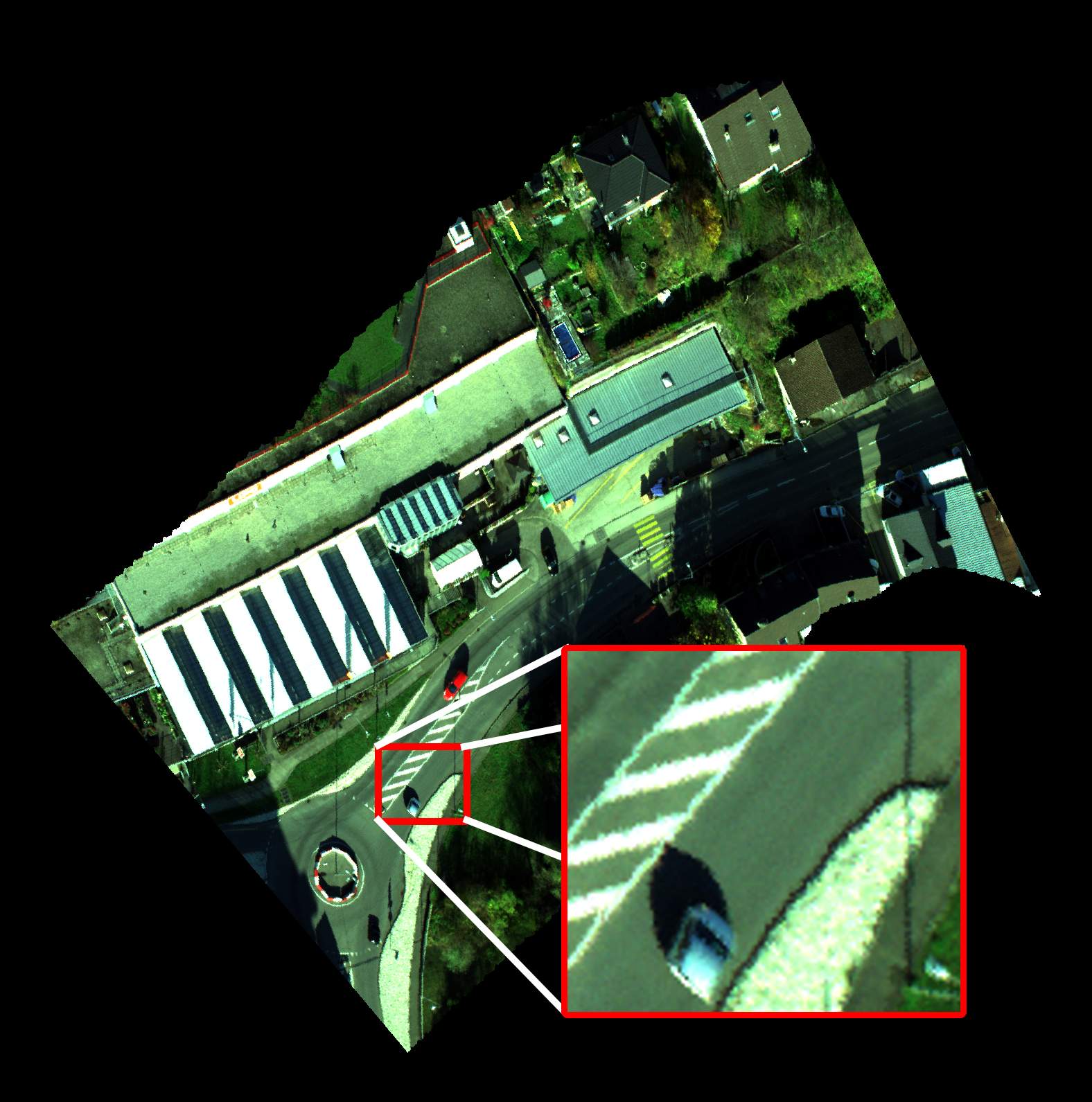}
        \subcaption{With time synchronization}
        \label{fig:time_sync_results:with}
    \end{subfigure}
    
    \caption{Visual quality estimation of rectified hyperspectral image using GPS/INS (a) without and (b) with time synchronization.}
    \label{fig:time_sync_results}
\end{figure}

\noindent The quality of the results can easily be verified visually, as shown in Figure~\ref{fig:time_sync_results}. The georeferenced image quality drastically increases with time synchronization without parasite waves pattern.

This shows that, our method for horizontal rectification can not only be used to generate high quality tie points, but can also be applied to estimate time offsets between trajectory and image data. This is useful for cheap and lightweight systems where time synchronisation is prone to error. For advanced systems, time tagging is typically implemented with hardware pulse events, where the number of events should match the number of lines registered by the hyperspectral camera (which was not the case for the Pika-L for unknown reasons).

\section{Solving the maximum a-posteriori estimate for our Bayesian model for horizontal shift estimation}

As presented in the paper, our probabilistic model for $\Delta x$ and $\Delta y$ is:

\begin{equation}
     \Delta \hat{x}, \Delta \hat{y} = \argmax_{\Delta x,\Delta y} p(\Delta x,\Delta y|\vec{I}) = \argmax_{\Delta x,\Delta y} p(\vec{I}|\Delta x,\Delta y)p(\Delta x)p(\Delta y).
 \end{equation}

 \noindent which we can solve by minimizing the negative log likelihood instead. An extremum is found when:
\begin{equation}
\label{eq:map_problem}
    \nabla (\log p(\Delta x,\Delta y|\vec{I})) = \nabla \log p(\vec{I}|\Delta x,\Delta y) + \nabla \log p(\Delta x) + \nabla \log p(\Delta y) = 0. 
\end{equation}

\noindent We use l-BGFS \cite{liu1989limited} to solve Equation~\ref{eq:map_problem}, which, unlike Newton method, does not require access to the Hessian, only the gradient. 

The derivatives of the prior are trivial:

\begin{equation}
\begin{split}
	\frac {\partial \log p(\Delta x)}{\partial \delta_x} & = 2\Delta x \\
	\frac {\partial \log p(\Delta y)}{\partial \delta_y} & = 2,
\end{split}
\end{equation}

\noindent and developing $\nabla \log p(\vec{I}|\Delta x,\Delta y)$ can be done using Matrix calculus:

\begin{equation}
\begin{split}
	\frac {\partial \log p(\vec{I}|\Delta x,\Delta y)}{\partial \Delta x} &= \sum \frac {\partial \log p(\vec{I}|\Delta x,\Delta y)}{\partial \matr{\Sigma}} \odot \frac {\partial \matr{\Sigma}}{\partial \Delta x}\\
	\frac {\partial \log p(\vec{I}|\Delta x,\Delta y)}{\partial \Delta y} &= \sum \frac {\partial \log p(\vec{I}|\Delta x,\Delta y)}{\partial \matr{\Sigma}} \odot \frac {\partial \matr{\Sigma}}{\partial \Delta y},
\end{split}
\end{equation}

\noindent with $\odot$ the element-wise product and $\sum$ being the sum of entries in the matrix. We have that:

\begin{equation}
\label{eq:impossible_equation}
	\frac {\partial \log p(\vec{I}|\Delta x,\Delta y)}{\partial \boldsymbol{\Sigma}} = \matr{\Sigma}^{-1} - \matr{\Sigma}^{-1} \vec{I} \vec{I}^\top \matr{\Sigma}^{-1},
\end{equation}

\noindent and $\frac {\partial \matr{\Sigma}}{\partial \Delta x}$ and $\frac {\partial \matr{\Sigma}}{\partial \Delta y}$ are given by the derivatives of the covariance kernel we are using.

\section{Estimation of the line patch size for horizontal displacement estimate}

\begin{figure}
    \centering
    \includegraphics[width=0.24\textwidth]{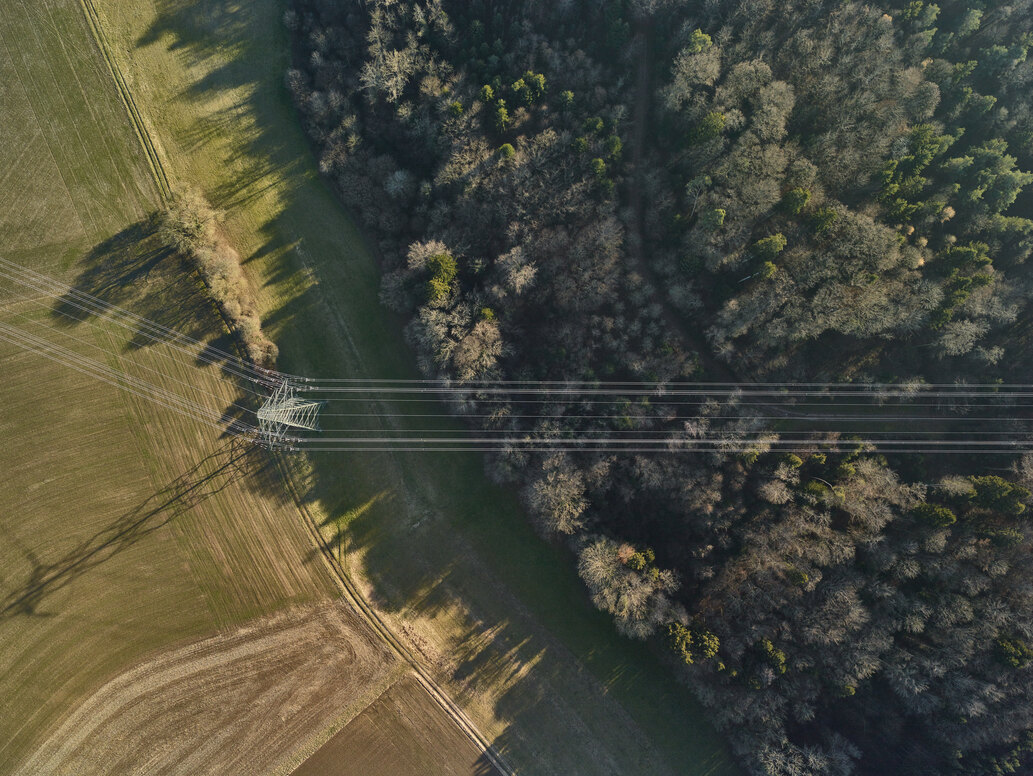} 
    \hfill
    \includegraphics[width=0.24\textwidth]{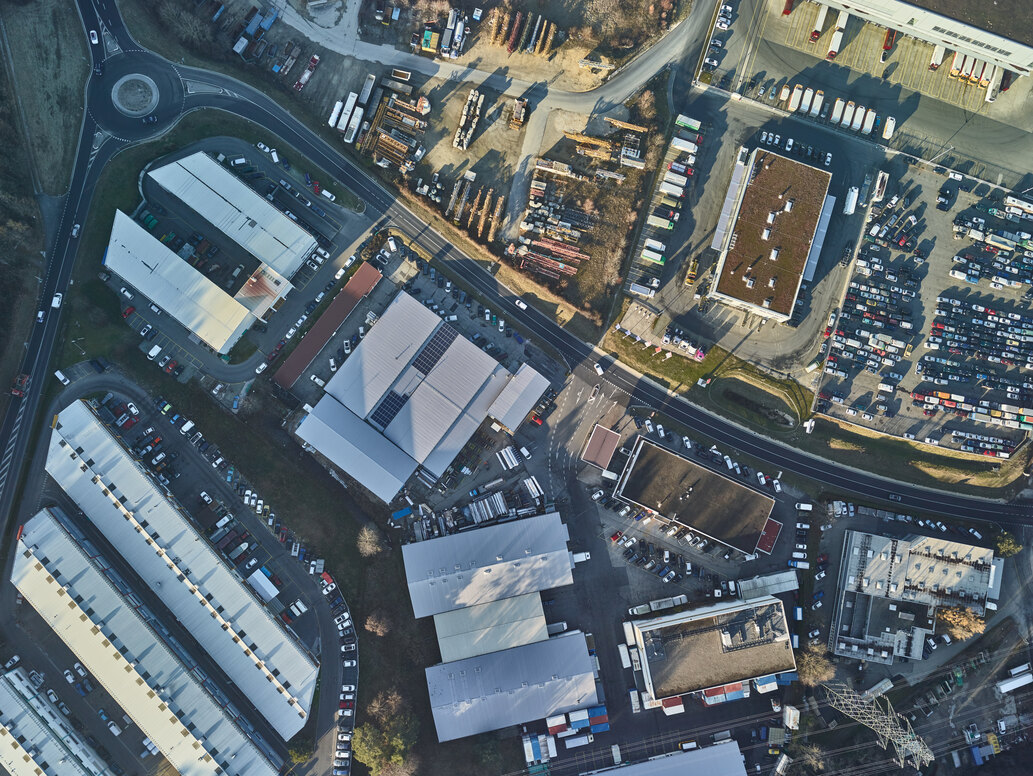} 
    \hfill
    \includegraphics[width=0.24\textwidth]{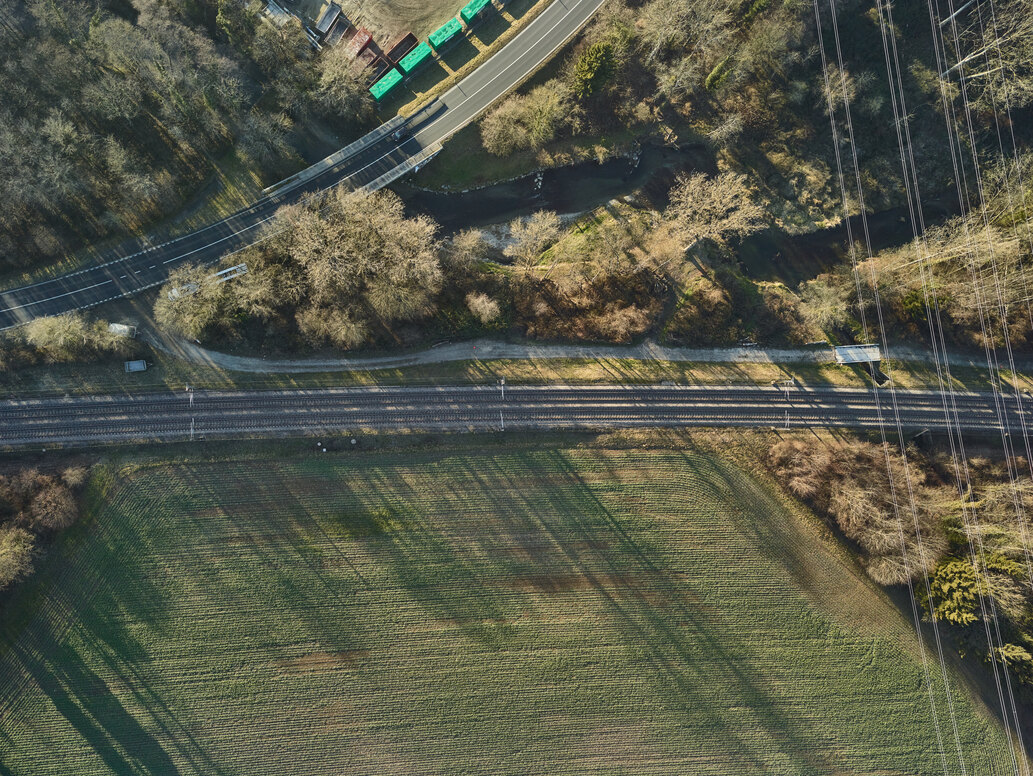} 
    \hfill
    \includegraphics[width=0.24\textwidth]{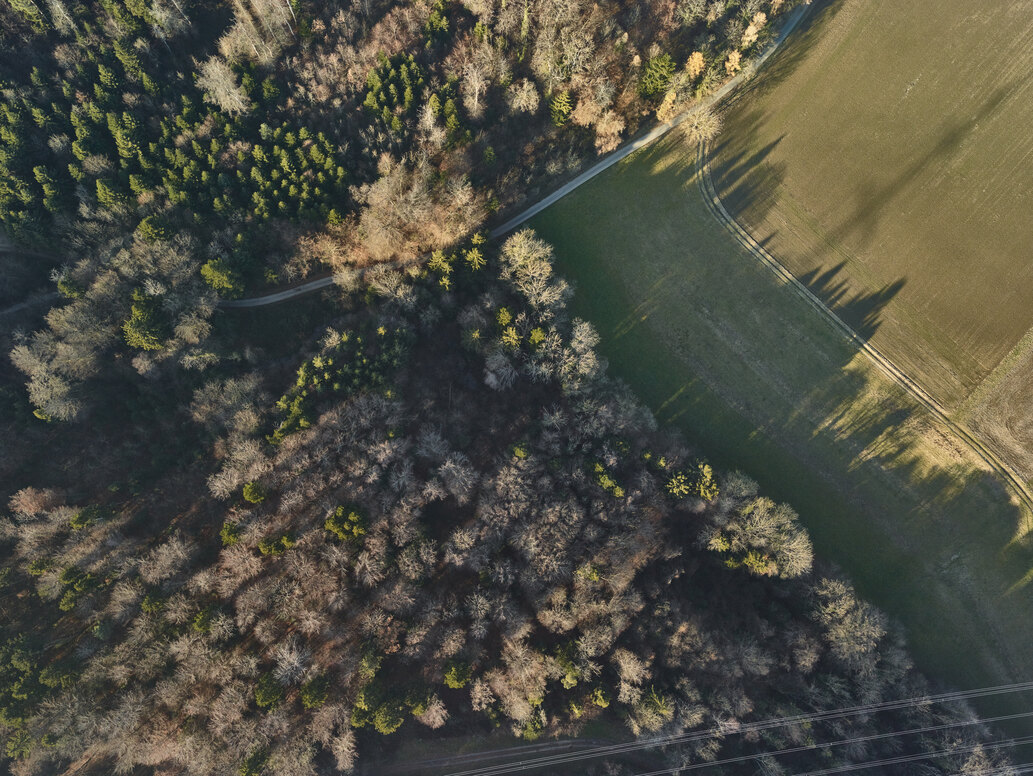}
    
    \caption{Example of Images we used to generate our simulated patches}
    \label{fig:simulated_patches_source}
\end{figure}

To estimate the best patch size for our Bayesian model, we use simulated lines with random shifts. To match plausible images, lines were extracted randomly from aerial frame imagery from an older flight we had available, see Figure~\ref{fig:simulated_patches_source}. We used a different flight than the one in Delemont to increase the variety in the data. To account for sub-pixel positions, we used cubic interpolation to resample the lines.

100 patches were generated for each shift, the absolute error was computed and each trial was repeated 100 times, generating a distribution of absolute errors.

We evaluated two common Gaussian process correlation kernels \cite{williams2006gaussian} for our method, the Exponential quadratic kernel and the Matérn kernel of order 3/2. 

The method of Fang \etal \cite{fang2017two} also uses sliding windows, but their work gives no value for the size of the sliding window used. To select the optimal window size for their method, we also evaluated the best correlation windows size for their method for 7 patch widths between 5 and 35 pixels.

\begin{figure}
    \centering
    \includegraphics[width=\textwidth]{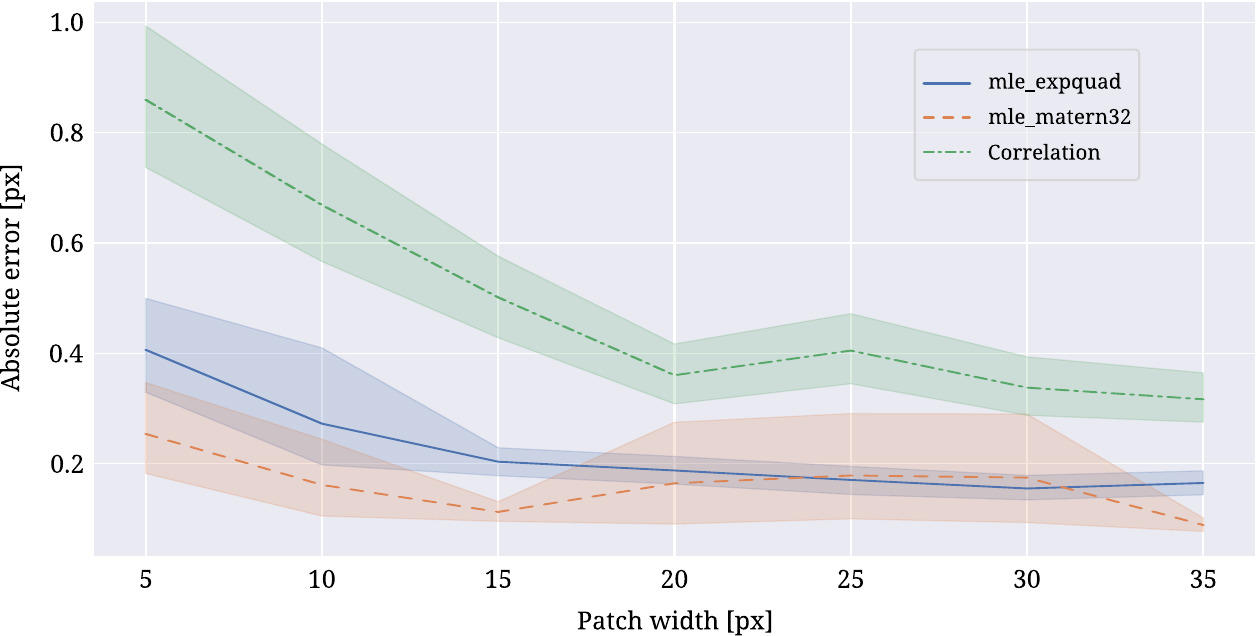}
    \caption{Effect of patch width on the accuracy of different kernel functions (and correlation) for horizontal shift estimation}
    \label{fig:patch_width_influence}
\end{figure}

The results, shown in Figure~\ref{fig:patch_width_influence}, show that all methods start by increasing their accuracy and then reach a plateau. Both Gaussian process Kernels reach their plateau at around 15 pixels, while the correlation based approach reaches it later, at 20 pixels. This delay indicates that our Bayesian model is more efficient at extracting information from fewer pixels. The Matérn kernel of order 3/2 seems to be the better option, albeit with a very small advantage over the Exponentiated Quadratic kernel. The correlation approach is not as accurate. 

Since the optimum patch size for our model seems to be around 15 pixels, we decided to use a patch size of 16 pixels, which is close, but also a power of two. Processing patches with size equal to powers of 2 is easier with SIMD instructions on modern processors.

\section{Horizontal shift correction on additional dataset}

We used a public dataset, published by Kim \etal, \cite{Kim_Alaska_UAV_Hyperspectral} to double check the accuracy of our horizontal rectification method.

\begin{figure}
	\centering
 
   	\includegraphics[width=0.8\textwidth]{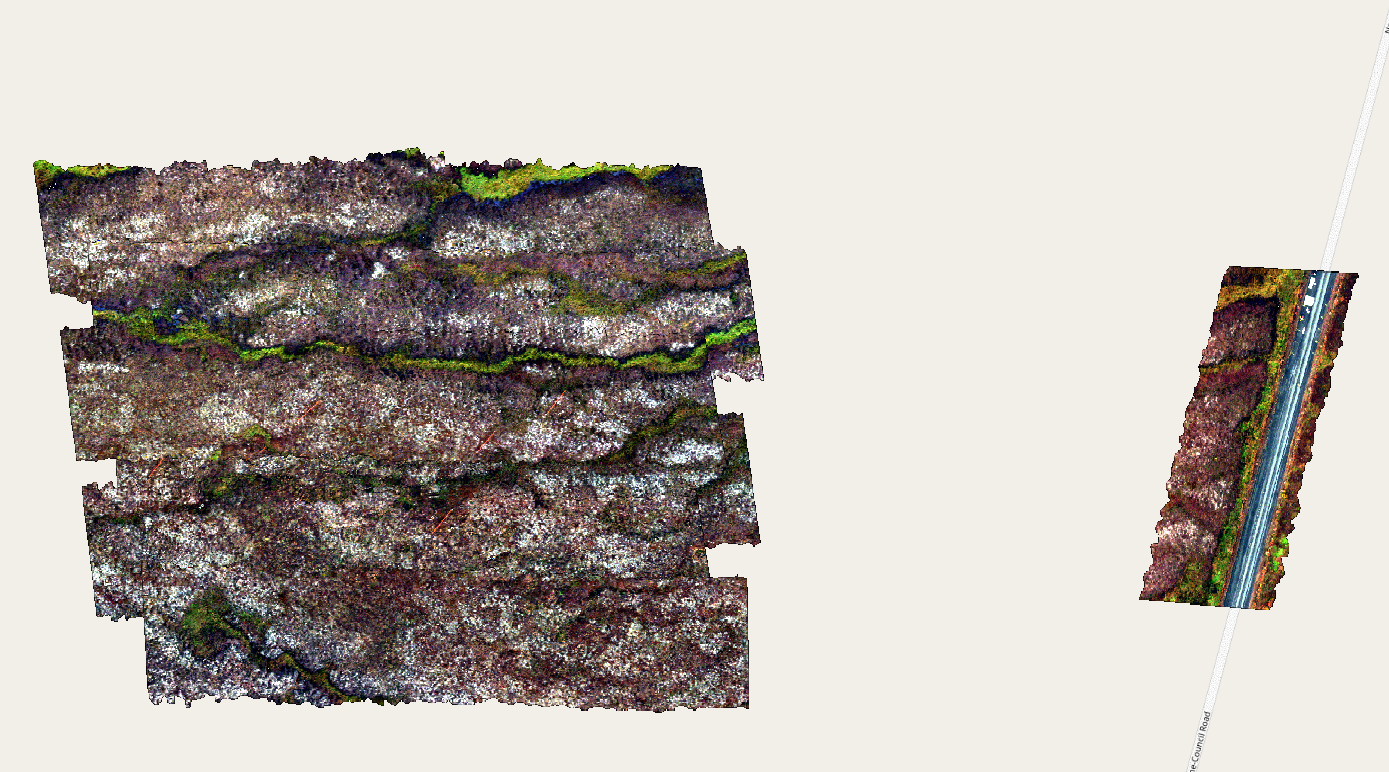}
			\caption{Calibration area and 6 overlapping flight segments in the dataset, georectified, map background \textcopyright~OpenStreetMap contributors}
   \label{fig:alaska_ortho_vnir}
	
\end{figure}

The dataset has been acquired over Council, Alaska, in 2019. One line has been acquired over the road with calibration targets and ground control points, then, 6 parallel overlapping lines have been scanned including different kinds of textures and vegetation, see Figure~\ref{fig:alaska_ortho_vnir}.

\begin{figure}
    \centering
    \includegraphics[width=0.6\textwidth]{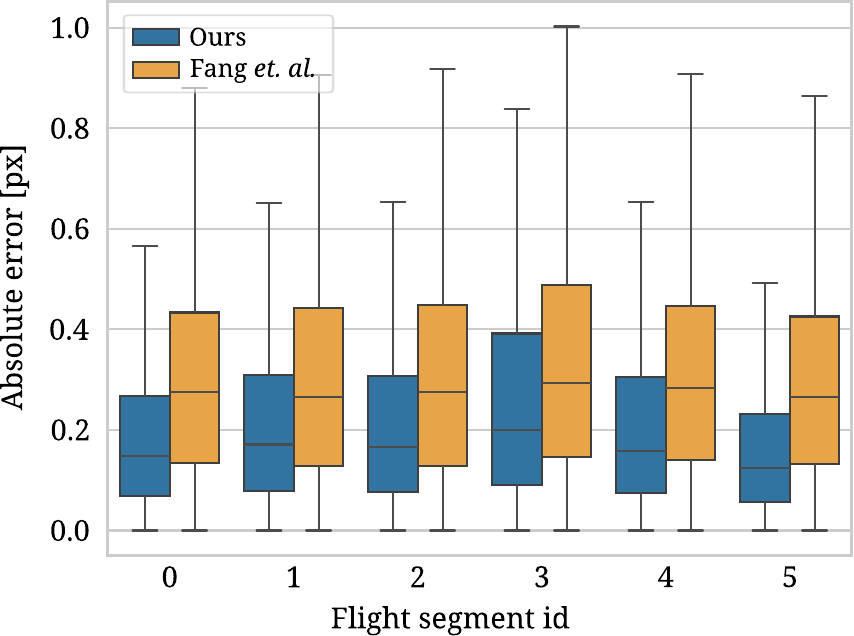}
    \caption{Distribution of horizontal shift corrections absolute errors for the Alaska dataset}
    \label{fig:alaska_errors}
\end{figure}

This dataset offers multiple benefits compared to the dataset we acquired over Delemont. First, it is publicly available at the time of writing this paper, while ours will be released only after an embargo period. Second, the 6 main lines in the dataset contain natural features, no strong visual features like roads or buildings. This dataset also has a major drawback though, which is that it contains no perpendicular intersection in the push-broom data. This means that the calibration of the Boresight using only tie points is highly under-constrained for the roll angle. As epipolar constraints do not restrict the motion along the axis parallel to the constraint, if all constraints are oriented in the same direction, the corresponding rotation axis cannot be determined.

Nonetheless, we evaluated our Bayesian model for horizontal rectification against the method of Fang \etal \cite{fang2017two} on this dataset. The methodology is the same as we applied to our dataset in the main paper. The results on each of the 6 lines, shown in Figure~\ref{fig:alaska_errors}, demonstrate that our method is able to generalize to different kinds of images, with different texture and overall appearance. The approach of Fang \etal \cite{fang2017two} also generalizes, and like in our dataset our method is slightly more precise. This is not unsurprising, as our model, as well as the one of Fang \etal \cite{fang2017two}, are based on very simple hypothesis that should be mostly valid for most data.

{
    \small
    \bibliographystyle{ieeenat_fullname}
    \bibliography{main}
}